\pdfoutput=1

\documentclass[11pt]{article}

\usepackage[]{EMNLP2023}

\usepackage{times}
\usepackage{latexsym}

\usepackage[T1]{fontenc}

\usepackage[utf8]{inputenc}

\usepackage{microtype}

\usepackage{inconsolata}

\usepackage{graphicx}

\usepackage{shortvrb}
\usepackage{amsmath,amssymb}
\usepackage{math_cmd}
\usepackage{xcolor}
\usepackage{algpseudocode}
\usepackage{algorithm}
\usepackage{breakcites}
\usepackage{pifont}
\usepackage{diagbox}
\usepackage{multirow}
\usepackage{booktabs}
\usepackage{enumitem}
\usepackage{bbm}
\usepackage{amsmath}
\usepackage{mathtools}

\usepackage{footnote}
\newcolumntype{L}[1]{>{\raggedright\let\newline\\\arraybackslash\hspace{0pt}}m{#1}}
\newcolumntype{C}[1]{>{\centering\let\newline\\\arraybackslash\hspace{0pt}}m{#1}}
\newcolumntype{R}[1]{>{\raggedleft\let\newline\\\arraybackslash\hspace{0pt}}m{#1}}

\title{Template-assisted Contrastive Learning of Task-oriented Dialogue Sentence Embeddings}

\author{Minsik Oh \\ Stanford University \\\And
  Jiwei Li \\ Zhejiang University \\\And
  Guoyin Wang \\ Alibaba Qwen Pilot \\\AND
  \textnormal{
  \texttt{minsik@stanford.edu},\quad
  \texttt{guoyin.wang@alibaba-inc.com}
  }
  }

\begin{document}
\maketitle
\begin{abstract}

Learning high quality sentence embeddings from dialogues has drawn increasing attentions as it is essential to solve a variety of dialogue-oriented tasks with low annotation cost. Annotating and gathering utterance relationships in conversations are difficult, while token-level annotations, \eg, entities, slots and templates, are much easier to obtain. Other sentence embedding methods are usually sentence-level self-supervised frameworks and cannot utilize token-level extra knowledge. We introduce \textbf{T}emplate-\textbf{a}ware \textbf{D}ialogue \textbf{S}entence \textbf{E}mbedding (TaDSE), a novel augmentation method that utilizes template information to learn utterance embeddings via self-supervised contrastive learning framework. We further enhance the effect with a synthetically augmented dataset that diversifies utterance-template association, in which slot-filling is a preliminary step. We evaluate TaDSE performance on five downstream benchmark dialogue datasets. The experiment results show that TaDSE achieves significant improvements over previous SOTA methods for dialogue. We further introduce a novel analytic instrument of semantic compression test, for which we discover a correlation with uniformity and alignment. Our code is available at \url{https://github.com/minsik-ai/Template-Contrastive-Embedding}.

\end{abstract}

\section{Introduction}
\label{sec:intro}

\begin{figure}[t]
\centering
\includegraphics[width=.9\columnwidth]{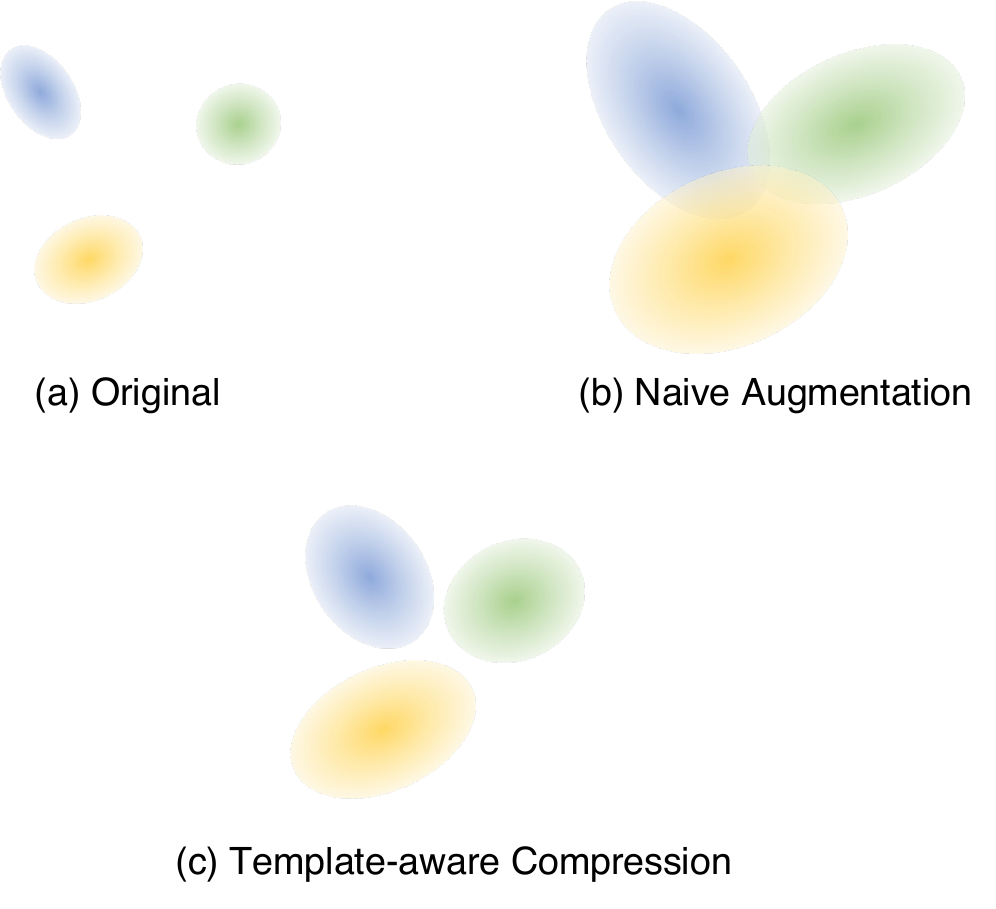}
\caption{Embedding hyperspace changes with our method, from (a), (b) to (c). Ellipses denote sentence representations from the dataset, belonging to unique semantic groups. (a) shows the limited original data, (b) shows the effect of noisy data augmentation in which semantic clusters overlap, and (c) shows enhanced semantic group separation with our methods, with templates within each semantic group to constrain the embeddings.}
\label{fig:nlu_repr}
\end{figure}


Learning sentence embeddings from dialogues has recently attracted increasing attentions~\cite{zhou2022learning, liu2021dialoguecse}.~\footnote{Different from general embeddings (Appendix~\ref{sec:eval}).} Learning high quality dialogue semantics~\cite{hou2020few,krone2020learning,yu2021few} helps solving various downstream tasks, especially in the scenarios with limited annotations~\cite{snell2017prototypical,vinyals2016matching,hyprank-2018, skill-routing-2021}. 

Contrastive Learning~\cite{contrast-2006} is a method to learn sentence embeddings by bringing semantically associated samples closer while pushing unrelated samples further apart. Unsupervised contrastive learning has been gaining momentum since it does not require human annotations, requiring supervision signals that augment original sentence. Some examples of supervision signals are document spans~\cite{declutr-2021}, Wikipedia entries~\cite{ease-2022}, consequent sentences~\cite{zhou2022learning}, prompt augmentations~\cite{promptbert-2022} and dropout hidden weights~\cite{simcse-2021}. 


Benefitting from the advance of contrastive learning, there has been solid success in learning universal sentence representations in both supervised~\cite{sbert-2019,feng-etal-2022-language} and unsupervised manner~\cite{simcse-2021,diffcse-2022,declutr-2021, ease-2022,promptbert-2022}. However, universal sentence embeddings usually achieve undesirable performance in dialogue domain~\cite{zhou2022learning,wu2020todbert}, since specific semantic relations between dialogue utterances exist (Appendix~\ref{sec:eval}).


In this paper, we explore how we can create semantically relevant sentence embeddings for dialogue. Templates~\cite{template-2020} and slots are high-quality auxiliary data for dialogue understanding purposes~\cite{hyprank-2018, slurp-2020, massive-2022}. They are a variable representation of text structure and salient slot values. However, previous sentence embedding frameworks cannot incorporate such information. We present TaDSE, \underline{T}emplate-\underline{a}ware \underline{D}ialogue \underline{S}entence \underline{E}mbedding generation framework which produces superior text embeddings for dialogue understanding via template-aware data augmentation, training, and inference.

Our template-based data augmentation method (Section~\ref{sec:data}) exploits salient ingredients already present in task-oriented dialogue - templates, entities (slots), and their values. General purpose data augmentation methods, \emph{e.g.}, rule-based methods or backtranslation~\cite{feng2021survey, wei2019eda, zhang2022virtual, qu2020coda, sennrich2016improving} are prone to semantic alterations or require a model~\cite{caseaugpos}. Our augmentation strategy produces consistently natural utterances and reinforces the dataset distribution in a realistic manner. We discover that our augmentation easily attain a stable performance increase, especially in combination with our training method, even if noise exists in synthetic data.

Our TaDSE training method (Section~\ref{sec:train}) encodes auxiliary template representations and their pairwise association with matching utterance representations. Each template is salient in regard to the semantic structure of the utterances, thus the model can improve itself by learning to distinguish between correct and mismatched utterance/template pairs. We introduce a pair of contrastive loss terms that designate the associated pairs of utterance and template as positive. Our pairwise training outperforms previous utterance-only unsupervised methods across five dialogue datasets.


Our TaDSE inference method (Section~\ref{sec:infer}), which we define as "semantic compression test", is an instrument to inspect another conjecture of our training method, an interpretation that bringing correct utterance and template representations closer enhances representation. By enhancing specific semantics in the templates, the model can differentiate cosmetically similar utterances. Semantic compression improves the performance on augmentation-stable datasets, in addition to a noteworthy correlation with existing tools of uniformity/alignment~\cite{uniformity-2020}.

Our contributions are summarized as follows:

\begin{enumerate}
    \item We propose a special synthetic data augmentation, a novel data augmentation approach that aims to replicate \textit{real-life} utterances. 

    \item We propose a novel training \& inference pairwise dialogue sentence embedding learning framework, justified via SOTA performances.

    \item Our experiments visibly show that the inferred utterance representations reshape the hyperspace in accordance with our expectations.
\end{enumerate}

\section{Related Works}
\label{sec:rel}

\begin{figure*}[t]
     \centering
     \includegraphics[width=1.7\columnwidth]{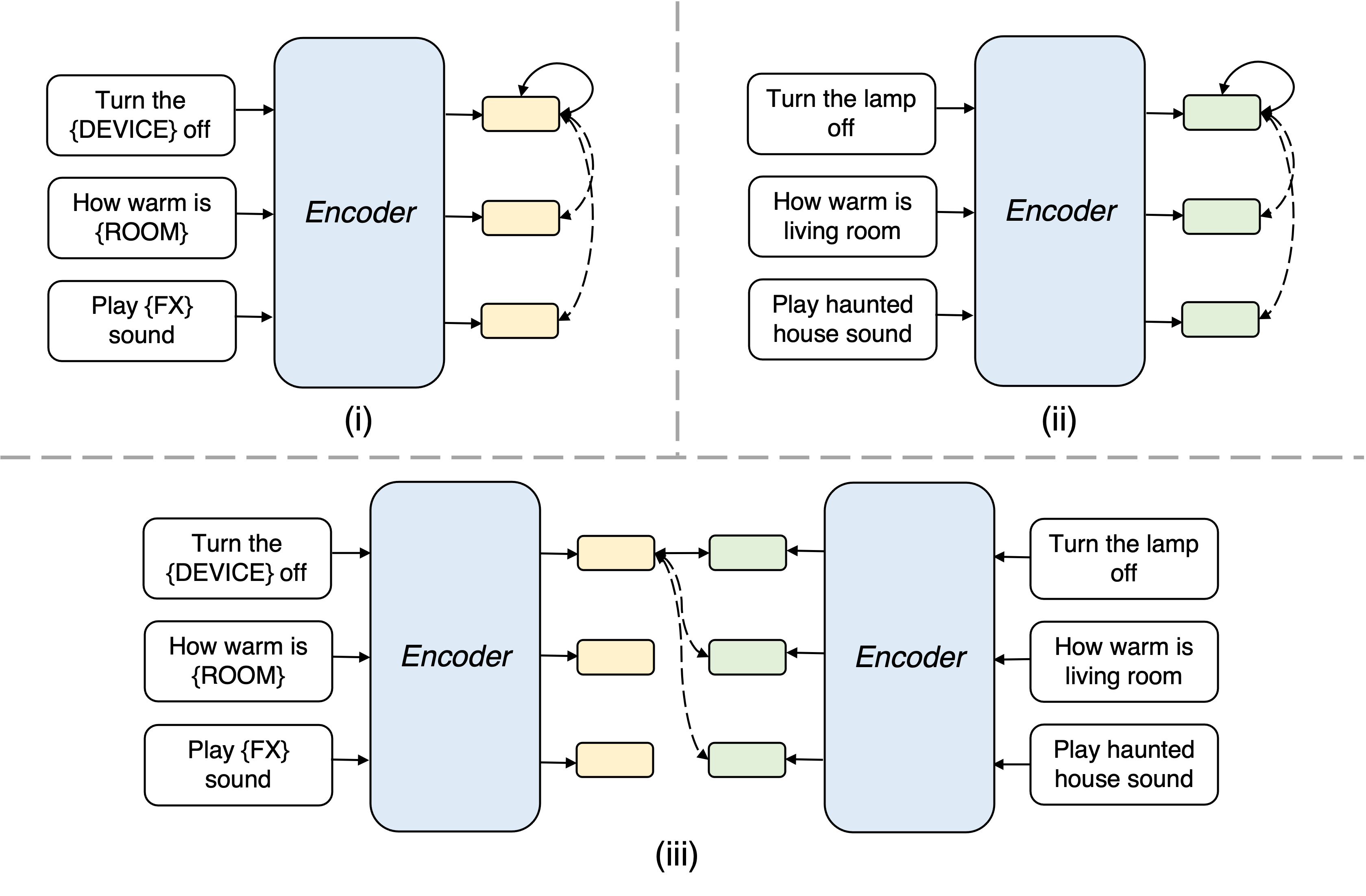}
     \caption{Our template contrastive learning methods. The first diagram displays template contrastive learning ($L^t$), second diagram displays utterance contrastive learning ($L^u$), and the third diagram displays pairwise contrastive learning ($L^\textit{pair}$). \textit{Encoder} represents the embedding generation model and yellow, and green represent template and utterance representations respectively. Solid bidirectional arrows designate positive pairs and dashed bidirectional arrows designate negative pairs.}
 \label{fig:contrast}
\end{figure*}

\begin{table*}[t]
    \centering
    \begin{tabular}{cccccccc}
        \toprule
        Src. Data & Strategy & Slots & Values & Templates & Orig. Utterances & Utterances & U/T Ratio \\
        \midrule
        SNIPS & top-5 & 39 & 11.9K & 7.4K & 13.1K & 163K & 22x \\
        ATIS & top-2 & 41 & 0.6K & 3.3K & 4.5K & 239K & 72x \\
        MASSIVE & top-3 & 55 & 4.0K & 10.3K & 11.5K & 78K & 8x \\
        HWU64 & top-3 & 56 & 6.0K & 16.6K & 19.2K & 133K & 7x \\
        CLINC150 & top-5 & 17 & 1.7K & 15.3K & 15.3K & 220K & 14x \\
        \midrule
        Total & - & 208 & 24.2K & 52.9K & 63.6K & 834K & 16x \\
        \bottomrule
    \end{tabular}
    \caption{Statistics of our augmented dialogue datasets. The "Slots" column is for slots ({DEVICE}, {ROOM}, etc.) while the "Values" column is values that fit the slots (television, lounge, etc.). "U/T ratio" denotes how many utterances exist per template on average after augmentation.}
    \label{tab:dataset_aug}
\end{table*}

\begin{figure*}[t]
     \centering
     \includegraphics[width=1.85\columnwidth]{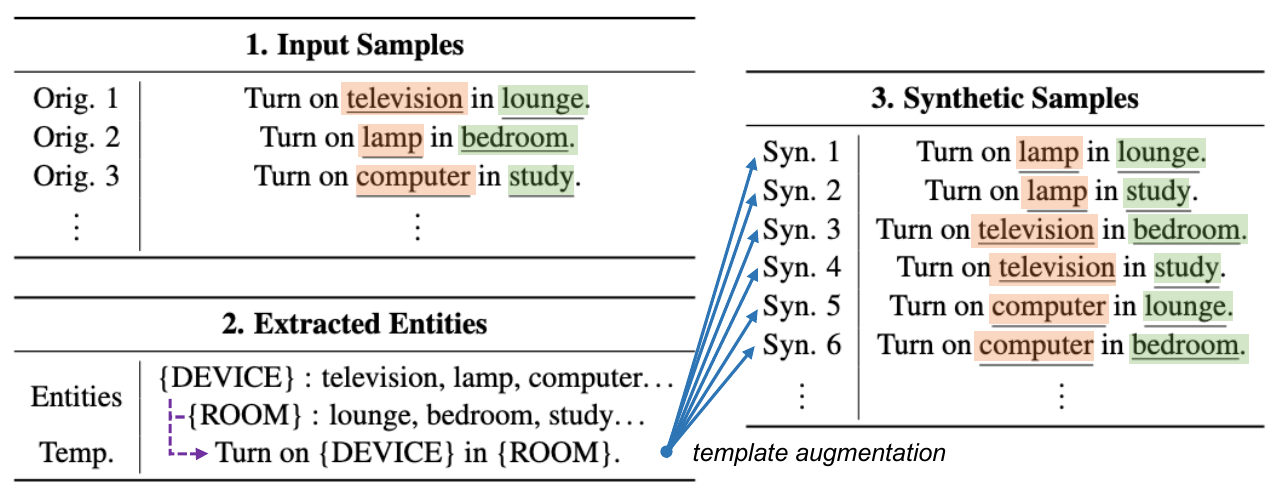}
     \caption{Our template data augmentation process in a simplified example, with a single template. In practice, thousands of templates and slot values exist per dataset (Table~\ref{tab:dataset_aug}). We experiment with both manual annotations and automated slot-filling method.}
 \label{fig:aug_sample}
\end{figure*}
Unsupervised Sentence Embedding methods train with contrastive objectives effectively to learn universal sentence embeddings. For vision, methods such as SimCLR~\cite{simclr-2020, simclrv2-2020} have demonstrated the importance of data augmentation in contrastive learning. In NLP, methods such as SimCSE, DiffCSE, PromptBERT~\cite{simcse-2021, diffcse-2022, promptbert-2022} show that simple augmentations such as dropout masking, token-wise masking, and prompt augmentation are an effective positive representation target. Our method differs from previously studied methods due to our novel application of semantically relevant token-wise templates for contrastive loss design.

Slot-filling and intent classification are major tasks for dialogue understanding purposes~\cite {survey}. Recent works perform the tasks jointly or in a multi-stage manner~\cite{slot1,slot2,slot3,slot4,slot5,slot6}. Each task has also been studied separately~\cite{survey, slu-rnn, slu-rnn2, slu-rnn3}. In line with prior works, we perform slot-filling as a necessary step for representation learning in the dialogue domain, after which we perform the intent classification task. We repurpose existing tasks to model the semantic structure of utterances and assess the quality of the semantic information. This brings the benefit of both conceptually sound methodology and practical NLU applications via enhanced embeddings.



Studying the representation space formed by learned embeddings has received influential attention, with recent work introducing uniformity/alignment to induce properties in relation to hypersphere~\cite{uniformity-2020}. Anisotropy problem is also identified with language representations~\cite{anistropy-word-2019, anistropy-sentence-2020, gao2018representation}, the problem of only narrow cone in the hyperspace being occupied by the embeddings. This behavior is also observed in multi-modal setting~\cite{multimodalrepr-2022}. While we utilize the hyperspace analysis tools provided by previous works, we introduce a novel instrument of semantic compression which has the marked benefit of being semantically interpretable in regards to the meaning of the natural language sentences.

\section{Proposed Method}
\label{sec:method}

\subsection{Template Data Augmentation}
\label{sec:data}

In dialogue datasets such as SNIPS~\cite{coucke2018snips} and ATIS~\cite{hemphill-etal-1990-atis}, multiple utterances correspond to a single template, which we express as "utterance-template pairwise association". We posit that strengthening the diversity of utterance-template pairwise associations is essential for our training scheme. This variety of utterances per template will be retained in distributions from actual \textit{real-life} scenarios.\footnote{Reasonably high percentage of customers booking an airplane ticket would tend to say "Could I book a plane ticket to \{CITY\}?" rather than complex variations of the template.} We present a template-based augmentation strategy to replicate realistic usage patterns, with the added benefit of providing varied natural utterances.

We select a set of slots (\textit{entities}) that are relevant to the dialogue domain, whether it be airlines, countries, or appliances, and categorize them to form a Slot Book. We construct permutations of the templates by filling the slot tokens with selected slot values. We select top-$k$ frequent slot values from the training set to maintain the quality of utterance-template association. This method (Fig.~\ref{fig:aug_sample}) may be extended further (Section~\ref{sec:limit}).

We utilize dialogue datasets of SNIPS~\cite{coucke2018snips}, ATIS~\cite{hemphill-etal-1990-atis}, MASSIVE~\cite{massive-2022}, HWU64~\cite{hwu64} and CLINC150~\cite{clinc150}. For MASSIVE, SNIPS, HWU64 and ATIS datasets, we utilize annotated templates and slot values already present in the dataset. For only CLINC150 dataset, we utilize a weak baseline of NER to automatically obtain slots and create templates, since no annotations are available. Slots of interest are cities, airlines, time, food, etc, with samples in Appendix~\ref{sec:clinc}. The CLINC150-specific configuration is intended to verify whether the noisy baseline slot-filling system is functional with our data augmentation, training, and inference framework. We leave enhanced slot-filling techniques for future work (Section~\ref{sec:limit}).

\subsection{Pairwise Modeling}
\label{sec:train}

\begin{figure*}[t]
     \centering
     \includegraphics[width=1.85\columnwidth]{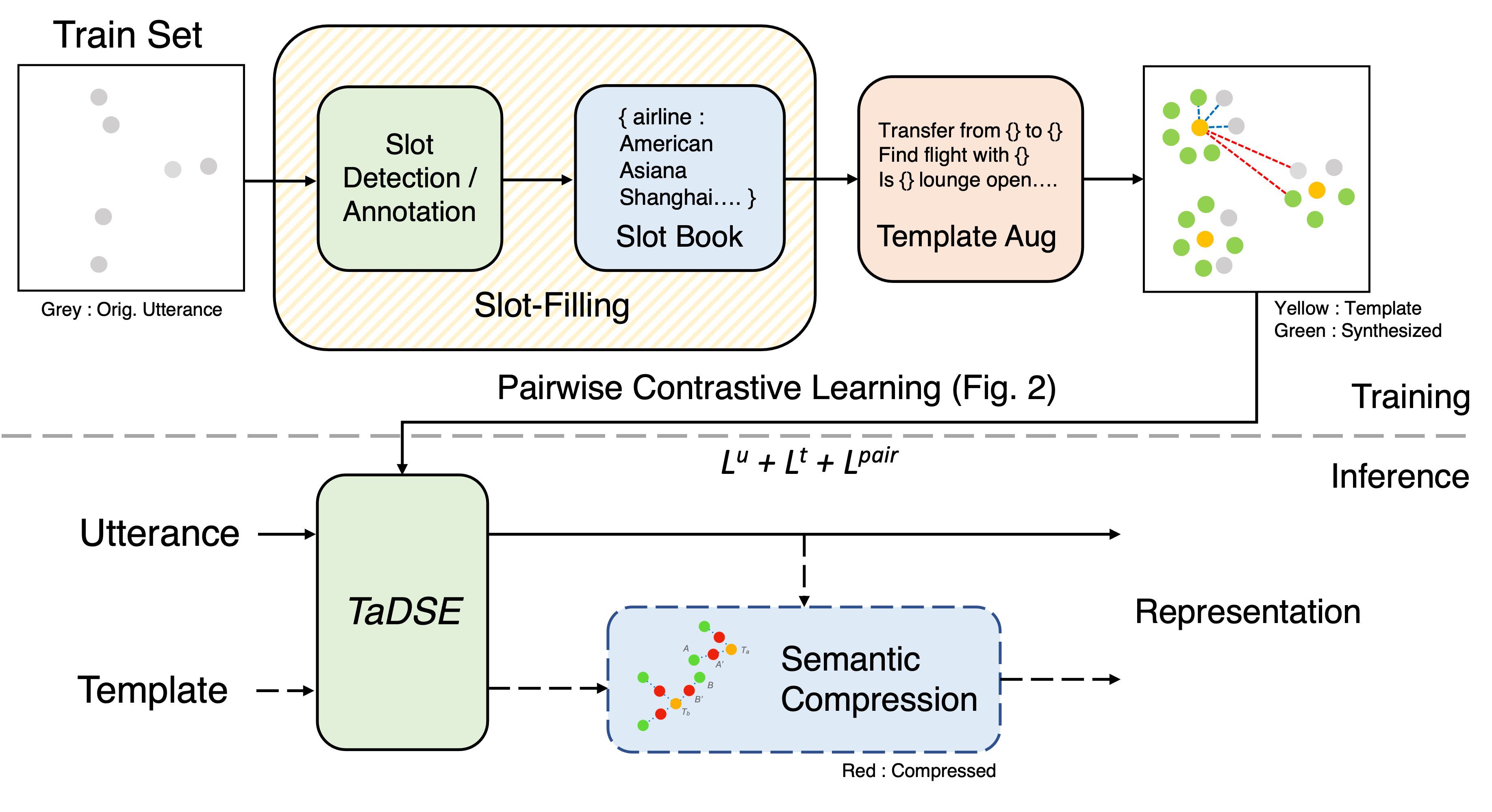}
     \caption{Our embedding generation process. Blue and red dashed lines are examples of positive and negative pairs for $L^{\textit{pair}}$ loss. Dashed arrows depict an alternative choice of semantic compression inference technique. Slot-filling baselines and template sources described in Section~\ref{sec:data}.}
 \label{fig:system}
\end{figure*}
The effect of "anchoring" the sentence representations with auxiliary data has been studied with NLI-reliant hard-negatives~\cite{simcse-2021}, Wikipedia entries in multi-lingual settings~\cite{ease-2022}, with document spans~\cite{declutr-2021}, and co-occurring utterances in pre-training dialogue corpus~\cite{zhou2022learning,liu2021dialoguecse,wu2020todbert}. The aforementioned studies rely on incidental auxiliary data or pre-trained auxiliary models and thus are heavily reliant on distributions in a large corpus. We introduce a concept of "pairwise anchoring", in which we train with an auxiliary template generated from the utterance itself via tokenwise masking. This benefits the training procedure since it is impossible to pair the sentence with irrelevant data, and requires only a small training set for fine-tuning. We teach the model the capability to distinguish correct utterance and template pairs via contrastive learning (Fig.~\ref{fig:contrast}). 

First, we define \textbf{template representation loss}, where we train a saliently masked anchor with which we further induce utterance representations. We train with contrastive loss and the data augmentation with dropout noise according to~\cite{simcse-2021} framework. Let $\text{sim}(t_i, t_j)$ be cosine similarity $\frac{t_i^{T} t_j}{||t_i||\cdot||t_j||}$. Template representation is given as $t_i$ and dropout-variant as $t^+_i$. Negative representations sampled from mini-batch are $t_j$. The template loss function is:

\begin{align}
    &L_i^{t} = - \log \frac{e^{\text{sim}(t_i, t^+_i)/\tau_{t}}}{\Sigma_{j=1}^N e^{\text{sim}(t_i, t_j)/\tau_{t}}}
    \label{eq:p_contrast}
\end{align}

where $\tau_{t}$ is temperature hyperparameter for template representation. In addition, we further experiment with a trainable MLP layer $W_A$ to modify template representation $t$ as $t'=W_{A} t$. This is an extension of pooling experiments performed on~\cite{simcse-2021}, the difference being that we focus on the effect of including MLP for templates only in an asymmetric configuration.\\
\bigskip

Next, we compute \textbf{utterance representation loss} similarly in a contrastive manner. This is to ensure we correctly learn utterance representation without over-reliance on templates. If we define $u_i$, $u_i^+$ and $u_j$ similarly to $t_i$, $t_i^+$ and $t_j$ in Eq.~\ref{eq:p_contrast}~\cite{simcse-2021}, the utterance loss function is:

\begin{equation}
    L_i^u = - \log \frac{e^{\text{sim}(u_i, u_i^+)/\tau_u}}{\Sigma_{j=1}^N e^{\text{sim}(u_i, u_j)/\tau_u}}
    \label{eq:u_contrast}
\end{equation}

where $\tau_u$ is temperature for the utterance representation.

Lastly, we introduce \textbf{pairwise representation loss}, where we distinguish between correct and negative utterance-template pairs via contrastive learning to teach how certain semantically similar representations should group together (Section~\ref{sec:infer_res}). We compare within utterances instead of templates as to ensure unique negatives in relation to the template augmented data in Section~\ref{sec:data}. Defining $t_i$, $u_i$, $u_j$ same as Eq.~\ref{eq:p_contrast},~\ref{eq:u_contrast}, (note we draw negative mini-batch representations from utterances), the pairwise loss function is defined as:

\begin{equation}
    L_i^\textit{pair} = - \log \frac{e^{\text{sim}(t_i, u_i)/\tau_\textit{pair}}}{\Sigma_{j=1}^N e^{\text{sim}(t_i, u_j)/\tau_\textit{pair}}}
    \label{eq:pair_contrast}
\end{equation}

where $\tau_\textit{pair}$ is temperature for the pairwise representation.

\bigskip
Finally, combining losses $L_i^{t}$, $L_i^{u}$, $L_i^\textit{pair}$ defined in Eq.~\ref{eq:p_contrast},~\ref{eq:u_contrast},~\ref{eq:pair_contrast}, our training loss is the following :

\begin{equation}
   L_i^\textit{train} = L_i^{t} + \lambda^{u}L_i^{u} + \lambda^\textit{pair}L_i^\textit{pair} 
   \label{eq:train_contrast}
\end{equation}

\bigskip
where $\lambda^{u}$, $\lambda^\textit{pair}$ are hyperparameters to scale the importance of utterance and pairwise learning.

\begin{table*}[t]
    \centering
    \begin{tabular}{cC{1.5cm}C{1.5cm}C{1.5cm}C{1.5cm}C{1.5cm}C{1.5cm}}
        \toprule
        Model Type & SNIPS & ATIS & MASS. & HWU64 & Clinc150 & Average \\
        \midrule
        BERT & 80.00 & 78.05 & 41.86 & 50.84 & 33.35 & 56.82 \\
        SimCSE & 91.71 & 85.67 & 76.77 & 81.08 & 71.00 & 81.25 \\
        SimCSE (ours) & 92.00 & 86.56 & 77.27 & 80.24 & 71.05 & 81.42\\
        TOD-BERT & 90.71 & 81.75 & 58.47 & 63.25 & 50.60 & 68.96 \\
        TOD-BERT (ours) & 91.00 & 81.63 & 59.92 & 61.33 & 51.11 & 69.00\\
        DSE & 95.86 & 87.01 & 76.77 & 79.28 & 70.16 & 81.82 \\
        DSE (ours) & 95.86 & 84.66 & 73.50 & 76.75 & 68.51 & 79.86 \\
        \textbf{TaDSE} & \textbf{97.00} & \textbf{89.70} & 78.18 & \textbf{82.77} & 70.56 & 83.64 \\ 
        \textbf{TaDSE w/ MLP} & 96.29 & 89.14 & \textbf{79.15} & 82.29 & \textbf{72.49} & \textbf{83.87} \\ 
        \bottomrule
    
    \end{tabular}
    \caption{Unsupervised sentence embedding performance on intent classification task. "ours" models are trained with our augmented training set. More comments on the evaluations with dialogue datasets are in Appendix~\ref{sec:eval}. Comparison with supervised black-box embeddings in Section~\ref{sec:supervised}.}
    \label{tab:main_exp}
\end{table*}

\subsection{Semantic Compression}
\label{sec:infer}

In addition to the training procedure in Section~\ref{sec:train}, we introduce a new modification for inference as an instrument to examine our hypothesis about the semantic correlation between templates and utterances. Specifically, we measure how much it is possible to compress the hyperspace towards superior performance in a semantically interpretable process. The optimal value of compression coefficient $\lambda^\textit{comp}$ denotes the semantic well-formedness of the representations.

Our inference method of semantic compression is as follows: rather than just producing utterance representation as an inferred result, we introduce a scaled template representation term. This method augments the performance of the model, in addition to functioning as a tool to find optimal $\lambda^\textit{comp}$ on separate validation set. This results in the explicit inclusion of salient anchor representation with the new representation form, via which we enhance specific semantics in the templates. We show the effect in Fig.~\ref{fig:system}. New compressed representation $\textit{repr}_i$ is as follows :

\begin{equation}
    \textit{repr}_i = \lambda^\textit{comp} t_i + (1-\lambda^\textit{comp}) u_i
    \label{eq:infer}
\end{equation}

where $\lambda^\textit{comp}$ is relative importance of template representation with range $0 \leq \lambda^\textit{comp} \leq 1$.
\section{Experimental Setup}
\label{sec:exp}

We experiment with transfer learning on top of SimCSE~\cite{simcse-2021} BERT-base model, as to influence expected TaDSE properties on a representation model. We utilize kNN with the training set to select relevant reference vectors (Fig.~\ref{fig:intent_acc}) and compute intent detection accuracy (Section~\ref{sec:rel}). Baselines of TOD-BERT~\cite{wu2020todbert} and DSE~\cite{zhou2022learning} are dialogue embedding models utilizing utterance-only contrastive learning.  We do not perform STS evaluation due to domain mismatch and lack of context-aware semantics (Appendix~\ref{sec:eval}). More details in Appendix~\ref{sec:config}.

\section{Results}

\subsection{Main Results}
\label{sec:main_res}

We report the results of unsupervised learning evaluation in Table~\ref{tab:main_exp} and Table~\ref{tab:pairwise_exp}. We discover that our models consistently outperform other unsupervised learning embeddings. In particular, we observe a 5 - 6\% performance increase for SNIPS and ATIS datasets over the baseline. This is in line with the observation regarding augmentation stability in Section~\ref{sec:aug_res}. In addition, we find that augmenting template representation with a trainable MLP layer achieves similar performance. This is in line with observation in~\cite{simcse-2021} where inference with or without MLP achieve comparable performance (more experiments in Table~\ref{tab:pairwise_exp}, Fig.~\ref{fig:epoch}).

\subsection{Augmentation Stability}
\label{sec:aug_res}

\begin{table}[t]
    \centering
    \begin{tabular}{ccccccc}
        \toprule
        Data & Loss & Orig. & top-3 & top-4 & top-5 \\
        \midrule
        \multirow{2}{*}{SNIPS} & $L^u$ & 91.71 & 93.29 & 93.00 & 93.29 \\
        & $L^\textit{pair}$ & 91.71 & 93.71 & 95.14 & \textbf{96.14} \\
        \midrule
        \multirow{2}{*}{ATIS} & $L^u$ & 85.67 & 86.00 & N/A & N/A \\
        & $L^\textit{pair}$ & 85.55 & \textbf{89.59} & N/A & N/A \\
        \midrule
        \multirow{2}{*}{MASS.} & $L^u$ & 77.00  & 77.37 & 77.23 & 76.36 \\
        & $L^\textit{pair}$ & 77.30 & \textbf{79.39} & 79.29 & 78.41 \\
        \midrule
        \multirow{2}{*}{CLINC} & $L^u$ & 71.05 & 70.98 & 70.47 & 69.62 \\
        & $L^\textit{pair}$ & 71.27 & 72.25 & 72.73 & \textbf{72.98} \\
        \bottomrule
    
    \end{tabular}
    \caption{Template augmentation performance with single-source data. Note that ATIS reports top-2 instead of top-3, as we do not perform augmentations of higher order due to a large utterance count (Table~\ref{tab:dataset_aug}).}
    \label{tab:aug_exp}
\end{table}

We experiment with increased $k$ value in regards to the template-based augmentation process described in Section~\ref{sec:data} (Table~\ref{tab:aug_exp}). Each source datasets exhibit different characteristics in regard to augmentation - for example, the performance of SNIPS, ATIS models increases substantially with the higher order of augmentation (augmentation-stable), while MASSIVE models decrease after $3$. We detail this behavior in terms of stability regarding slot augmentations - while augmenting the templates with different entities assists in creating new salient utterances, the process may be compromised if sample-specific slot values are configured in the non-relevant templates. Thus, we assert that template and slot quality is important for token-based augmentation methods.

An interesting observation here is CLINC150 dataset, which we augment via a simple NER-based automatic slot-filling method instead of manual annotations. The process results in a highly noisy Slot Book specific to the dataset as described in Appendix~\ref{sec:clinc}, thus as expected the baseline utterance-only performance drops with higher-order augmentations. Interestingly, in contrast, $L^\textit{pair}$ models seem augmentation stable. This outcome inspires us to independently judge \textit{"slot correctness"}\footnote{Criteria for slot correctness would be: How granular are slots? Are right values assigned to correct slots?} and \textit{"template quality"}\footnote{Criteria for template quality would be: How many natural utterances would share templates? Are all salient entities identified and replaced?} - TaDSE is able to consider template quality in addition to slot correctness, while the utterance-only method would be greatly affected by low slot correctness and subsequent unnatural utterances. We leave further quantification of the observed behavior to future work.


\subsection{Pairwise Training}
\label{sec:train_res}

To study the effect of different losses introduced in Section~\ref{sec:train}, we perform experiments with single-source augmented datasets and report ablation results per selected losses (Table~\ref{tab:pairwise_exp}). Interestingly, the inclusion of template loss itself enhances the performance of the representations, showing the importance of salient semantic information stored in templates. The inclusion of pairwise loss further enhances performance, showing that training the models to distinguish utterance-template pairs enables models to learn superior representations.

We emphasize how augmenting templates with plausible utterance values unlocks TaDSE training, as augmented synthetic data increases utterances per template. The extra utterance-template pairs assist in the learning of discrimination capability. Results in Table~\ref{tab:aug_exp} show that the performance gap between TaDSE and the baseline method appears consistently with higher-order data augmentations.

    

\begin{table}[t]
    \centering
    \begin{tabular}{ccccc}
        \toprule
        Model & SNIPS & ATIS & MASS. & CLINC \\
        \midrule
        w/o aug & 91.71 & 85.67 & 77.00 & 71.05 \\
        aug & 93.29 & 86.00 & 77.37 & 70.98 \\
        $+L^{t}$ & 95.29 & 88.47 & 78.58 & 71.53 \\
        $+L^{t}$, $L^{\textit{pair}}$ & 96.14 & \textbf{89.59} & 79.39 & 72.98 \\
        $+L^{t'}$, $L^{\textit{pair}}$ & \textbf{97.00} & 88.69  & \textbf{79.83} & \textbf{73.45} \\
        \bottomrule
    
    \end{tabular}
    \caption{Pairwise training experiments for single-source models. The baseline is non-augmented original data trained via SimCSE method ($L^{u}$). Other models are trained with our augmented data, with model column depicting new losses.}
    \label{tab:pairwise_exp}
\end{table}

\begin{table}[t]
    \centering
    \begin{tabular}{ccccc}
        \toprule
        Model & SNIPS & ATIS & MASS. & CLINC \\
        \midrule
        $+L^{t}$ & 95.29 & 88.47 & 78.58 & 71.53 \\
        S.Comp. & \textbf{95.86} & 88.47 & 77.61 & \textbf{71.62} \\ 
        \midrule
        $+L^{t}$, $L^{\textit{pair}}$ & 96.14 & 89.57 & 79.39 & 72.98 \\
        S.Comp. & \textbf{96.43} & \textbf{90.03} & 78.24 & 72.25 \\ 
        \midrule
        $+L^{t'}$, $L^{\textit{pair}}$ & 97.00 & 88.69 & 79.83 & 73.45 \\
        S.Comp. & \textbf{97.29} & \textbf{89.36} & 77.47 & 72.71 \\ 
        \bottomrule
    \end{tabular}
    \caption{Semantic compression (S.Comp.) test for single-source data models, with model column depicting new losses with regards to $L^u$. The test succeeds on augmentation-stable datasets of SNIPS and ATIS.}
    \label{tab:inf_scale}
\end{table}

\subsection{Semantic Structure}
\label{sec:infer_res}

\begin{figure*}[t]
    \centering
    \includegraphics[width=1.8\columnwidth]{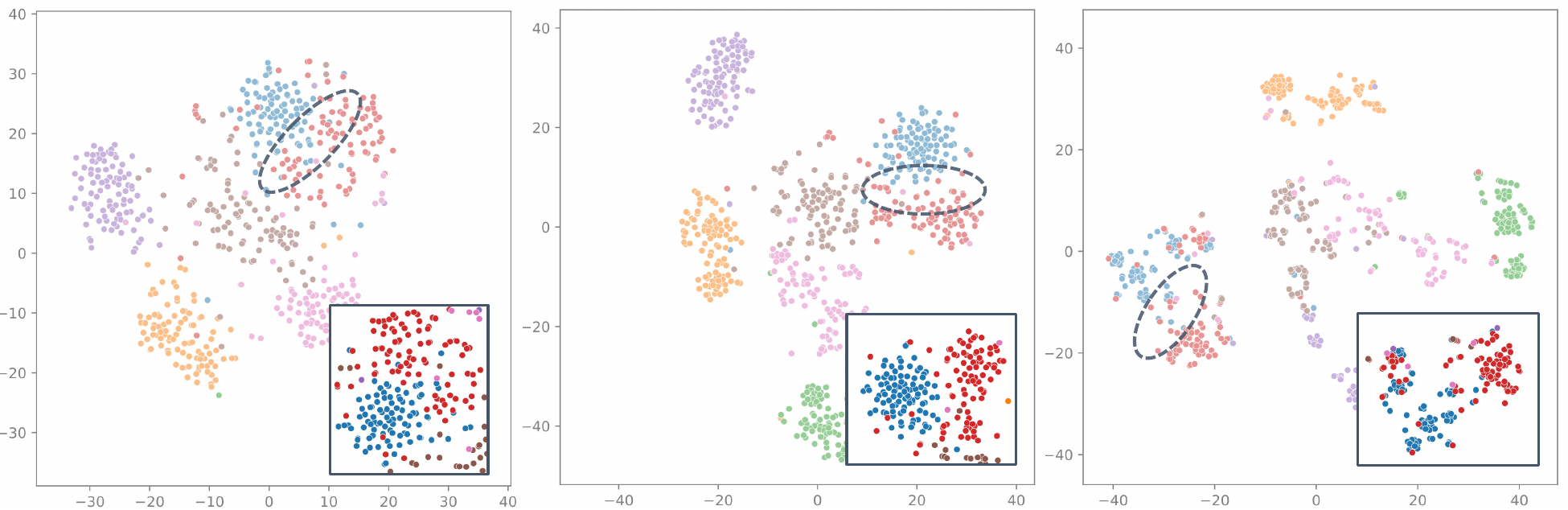}
    \caption{T-SNE diagram for SNIPS models, left : SimCSE, middle : TaDSE, right : TaDSE-compressed $0.5$. Embeddings are color-coded according to their labels, with red, blue colored embeddings being representations with PlayMusic, AddToPlaylist labels. We circle the increased sparcity near the effective decision boundaries and show a magnified view at lower right. Note that more compression does not always result in better performance. ATIS diagrams in Fig.~\ref{fig:atis_base},~\ref{fig:atis_train},~\ref{fig:atis_optimal}.}
    \label{fig:tsne_snips}
\end{figure*}

\begin{figure}[t]
    \centering
    \includegraphics[width=.8\columnwidth]{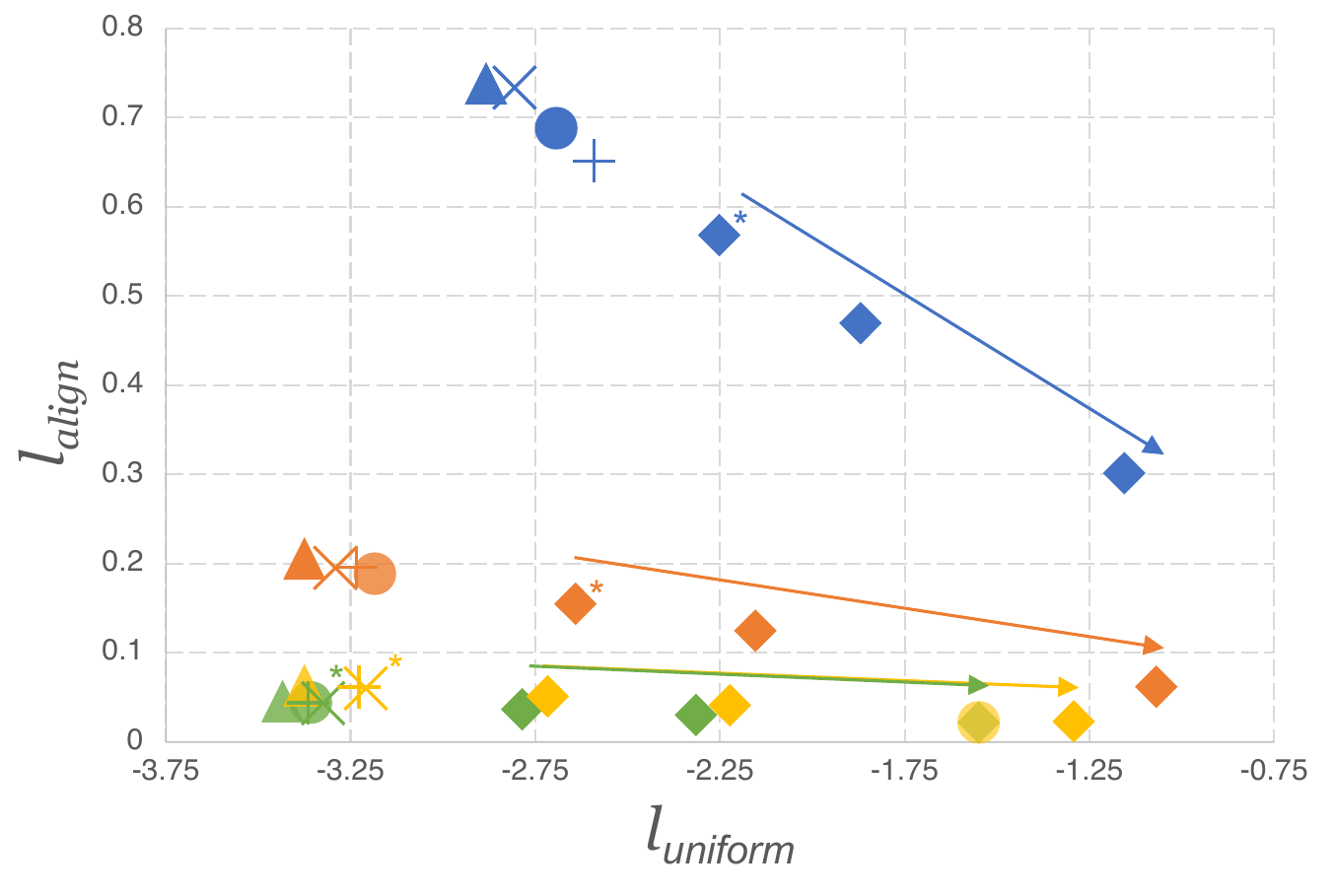}
    \caption{Uniformity / Alignment plot. Blue, orange, green, yellow are each ATIS, SNIPS, MASSIVE, CLINC models. Models for symbols are $+$ : SimCSE, $\times$ : TaDSE w/ MLP, $\triangle$ : utterance-only ($L^{u}$), $\circ$ : TaDSE, $\diamond$ : TaDSE-compressed. The arrows depict increasing $\lambda^\textit{comp}$. Most performant models marked with an asterisk (*). Lower values are superior.}
    \label{fig:uniform}
\end{figure}

We propose a semantic structure interpretation of the experimental results presented in Section~\ref{sec:train_res}. We assert that pairwise representation loss (Eq.~\ref{eq:pair_contrast}) brings utterance and template representations closer, enhancing semantic distances within utterance sub-cluster correlated with a template.

To examine the hypothesis, our semantic compression test (Section~\ref{sec:infer}) estimates how much we can enhance the aforementioned cluster properties by adjusting the hyperspace in a semantically interpretable way. Table~\ref{tab:inf_scale} reports that the test succeeds with augmentation-stable (Section~\ref{sec:aug_res}) datasets of SNIPS and ATIS\footnote{The test succeeds with $\lambda^\textit{comp} = 0.1\text{ or }0.2$, and $\lambda^\textit{comp} = 0.5$ is not selected for any of datasets. We leave experiments with continuous $\lambda^\textit{comp}$ values to future work.}, while non-stable datasets of MASSIVE and CLINC150 yield inconclusive results. This outcome shows the value of semantic structure interpretation and augmentation stability.



\subsection{Supervised Embeddings}
\label{sec:supervised}
 
We compare TaDSE with state-of-the-art commercial black-box and open-source embedding models (Table~\ref{tab:supervised}) to contextualize our results beyond dialogue-specific baselines. We evaluate OpenAI \verb|text-embedding-3-small| and \verb|text-embedding-3-large|, Google \verb|gemini-embedding-001|~\cite{lee2025geminiembeddinggeneralizableembeddings}, and Qwen \verb|Qwen3-Embedding-0.6B|~\cite{qwen3embedding} on the SNIPS and ATIS benchmarks with dimension size of 768.
Importantly, these models are known or likely to be \textit{supervised} sentence embeddings trained on large-scale similarity-labeled datasets, in contrast to TaDSE which requires no supervision labels.
 
\begin{table}[t]
\centering
\begin{tabular}{lccc}
\toprule
\textbf{Model} & \textbf{SNIPS} & \textbf{ATIS} & \textbf{Avg} \\
\midrule
Qwen3-0.6B & 92.14 & 89.14 & 90.64 \\
OpenAI-small & 97.86 & 84.88 & 91.37 \\
OpenAI-large & \textbf{98.57} & 84.77 & 91.67 \\
Gemini-001 & 98.29 & 86.00 & 92.15 \\
TaDSE & 97.00 & \textbf{89.70} & \textbf{93.35} \\
\bottomrule
\end{tabular}
\caption{Comparison with supervised black-box and open-source embeddings. Only TaDSE embeddings are unsupervised.}
\label{tab:supervised}
\end{table}
 
TaDSE achieves the highest average accuracy despite being the smallest model (110M parameters) and requiring no supervision labels, outperforming the best commercial black-box and open-source embedding models.
 
The per-dataset breakdown reveals a consistent pattern: commercial embeddings achieve near-ceiling performance on SNIPS, which contains only 7 intent classes (Appendix~\ref{sec:config} Table~\ref{tab:dataset}) and relatively simple utterance structures, while TaDSE leads on ATIS by a substantial margin over the commercial baselines.
ATIS utterances are characterized by compositional queries with complex syntactic structures.\footnote{e.g.

1. \textit{find me the earliest boston departure for atlanta and the lastest return trip from atlanta so that i can be in atlanta the longest amount of time but return to boston the same day}

2. \textit{show me all flights from pittsburgh to boston both direct and connecting that depart pittsburgh after 7 pm}

} We hypothesize that templates capture the compositional skeleton of such utterances in a way that surface-level supervised similarity training does not.
This aligns with our augmentation stability findings (Section~\ref{sec:aug_res}): both SNIPS and ATIS are augmentation-stable, yet ATIS is most sensitive to template-aware modeling, only requiring limited top-2 template augmentation compared to top-5 required by SNIPS to achieve similar performance increase of 5 - 6 \%. This may be because its structural complexity enables richer semantic anchoring.

TaDSE operates at 110M parameters, approximately $5\times$ smaller than Qwen3-Embedding-0.6B and orders of magnitude smaller than the infrastructure behind the OpenAI and Gemini embedding APIs.
The result suggests that domain-specific structural priors of utterance-template pairwise associations can effectively substitute for large-scale supervised training data and model capacity when the target domain exhibits complex compositional patterns.

\section{Analysis}

\subsection{Uniformity / Alignment}
\label{sec:uniform}

To identify inner workings of our methods, we utilize uniformity and alignment~\cite{uniformity-2020} to uncover how semantic structure of representations is altered. Definitions in Appendix~\ref{sec:uniform_align_def}.

We display uniformity/alignment for our models in Fig.~\ref{fig:uniform}. Surprisingly, we find that uniformity and alignment for TaDSE models have an inverse correlation.\footnote{This is a viable outcome considering that both uniformity and alignment are asymptotic of the same order to $\Vert f(a)-f(b) \Vert_2^2$, with distinct eligible representation pairs $(a, b)$. \cite{simcse-2021} report similar trends with certain variations.} We also find that utterance-only ($L^{u}$) models have the worst alignment, which relates to the naive augmentation step (Fig.~\ref{fig:nlu_repr}). Consequently, we report superior alignment and inferior uniformity for TaDSE models. This trade-off suggests that the performance increase may relate to superior alignment, especially in augmentation-stable (Section~\ref{sec:aug_res}) datasets of SNIPS and ATIS.

Importantly, we identify that TaDSE models from semantic compression test (Section~\ref{sec:infer}) obtain superior alignment correlated with higher $\lambda^\textit{comp}$ values. The results support our semantic structure interpretation in Section~\ref{sec:infer_res}. We conclude that our semantic compression test correlate with existing tools of uniformity/alignment.


\subsection{Qualitative Analysis}
\label{sec:sample_dist}

We graph a set of T-SNE diagrams\footnote{Per default Scikit-learn configuration of 30.0 perplexity.} for our representations (Fig.~\ref{fig:tsne_snips},~\ref{fig:atis_base},~\ref{fig:atis_train},~\ref{fig:atis_optimal}). We observe a clearer separation between music-associated clusters and a set of pronounced sub-clusters that correspond to semantic structure interpretation (Section~\ref{sec:infer_res},~\ref{sec:uniform}). 

    

\section{Conclusions}

In this work, we propose TaDSE, a novel unsupervised representation learning method that produces high-quality semantic representations of task-oriented dialogue. We develop a template-based data augmentation strategy that synthetically supplies diverse utterance-template pairs. We present methods of learning utterance-template discriminative capability and pairwise association via a new training scheme. We further justify the inner workings of our methods by applying a novel inference instrument that aligns well with uniformity/alignment analysis and visualization of representations. Our paper is first to employ semantic information in dialogue templates toward dialogue embeddings. Our template-aware data augmentation strategy and training losses enrich any dialogue datasets' semantic content. We believe that TaDSE is a reinforced text encoder for dialogue system applications.






\section{Limitations}
\label{sec:limit}

An enhanced slot-filling method for Clinc150 could be applied, rather than the current functional baseline method (the other 4 datasets use provided annotations). For example, slot-filling could suffer from a disambiguation problem, of slot values in different dialogue scenarios classified as the same slots. We leave enhanced slot-filling method experiments to future work. However, we emphasize that our methods work with automatically generated noisy Slot Book on Clinc150 dataset, which we describe in detail (Section~\ref{sec:aug_res}) in terms of \textit{template quality}. It will be interesting to observe how other slot-filling methods and hand annotations differ in terms of representation quality.

Our methods, in line with existing literature~\cite{zhou2022learning, wu2020todbert}, works with dialogue datasets of SNIPS~\cite{coucke2018snips}, ATIS~\cite{hemphill-etal-1990-atis}, MASSIVE~\cite{massive-2022}, HWU64~\cite{hwu64} and CLINC150~\cite{clinc150}. We actively avoid general datasets and evaluations such as STS (Appendix~\ref{sec:eval}). Future works may take inspiration from recent works in general embeddings~\cite{zhang-etal-2023-contrastive-learning, cheng-etal-2023-improving, wang2024text, wang2024improving, izacard2022unsupervised} and apply to dialogue domain. See also Appendix~\ref{sec:eval}.

The data augmentation process could be further improved with regard to the diversity of resulting utterances. Our baseline augmentation method improves performance (Table~\ref{tab:aug_exp}). However, the number of slot values for each slot is limited in the current setting. A smart slot value selection strategy such as one incorporating retrieval or randomization could be implemented in future works. 

While preprocessing model input\footnote{After data augmentation.}, we did not utilize distinct slot tokens to emphasize the semantic structure aspect. Instead, we replaced them as one token "\{SLOT\}" in templates. We leave it to future work to identify how discernible slot tokens in model inputs relate to representations.

It will be interesting to perform contrastive learning with token-level annotated "anchors" (Section~\ref{sec:train}) in other domains, such as legal or medical documents, and evaluate on such domains. Cross-domain evaluations can also be performed, with the model tuned on a domain that's different from selected evaluation dataset. We leave it to future work.

We experiment with discrete $\lambda^\textit{comp}$ values for semantic compression. While we observed sufficient trends for datasets with certain characteristics (Table~\ref{tab:inf_scale}, Fig.~\ref{fig:uniform}), we leave it to future work to perform semantic compression with continuous $\lambda^\textit{comp}$ values for follow-up evaluation. The test could also be theoretically expanded upon.

Our work only experiments with English, thus there is a potential risk of enhancing overexposure to the English language and its token-wise semantic characteristics, especially since we perform token-wise replacements and augmentations to the datasets. Templates that correspond to specific tokenization strategies might be necessary for experiments in other languages. 

\bibliography{anthology,custom}
\bibliographystyle{acl_natbib}

\newpage
\onecolumn
\appendix



\section{CLINC150 Slot Book}
\label{sec:clinc}

We only report top-5 occurrences for the allotted space. We experiment with the SpaCy NER model "\textit{en\_web\_core\_lg}" on the Clinc150 dataset. Please note that this is a noisy but effective baseline only for the Clinc150 dataset (discussion in Section~\ref{sec:limit}), required due to lack of annotations. The first table denotes slots. The second and 3rd tables are examples of well-formed slot books. The 4th and 5th tables are examples of noisy slot books - the 4th one is a combination of card companies, retirement funds, and bank names, and the 5th one is a combination of airlines, continents, and sightseeing locations. All categories should be separated for good \textit{slot correctness} (Section~\ref{sec:aug_res}). Note that even if the wrong slot is not in top-$k$ frequency, it still causes noisiness due to associated slots filled by incorrect values.

\begin{table}[H]
\begin{minipage}{.33\linewidth}
\centering
\begin{tabular}{cc}
\toprule
Slot & Count \\
\midrule
   GPE  &  1397\\
   DATE  & 1194\\
   ORG & 844 \\
   CARDINAL & 594 \\
   TIME & 443 \\
\bottomrule
\end{tabular}
\caption{Slots.}
\end{minipage}
\begin{minipage}{.33\linewidth}
\centering
\begin{tabular}{cc}
\toprule
Slot Value & Count \\
\midrule
   french  &  49\\
   italian  & 28\\
   spanish & 23 \\
   british & 23 \\
   mexican & 14 \\
\bottomrule
\end{tabular}
\caption{NORP slot values.}
\end{minipage}
\begin{minipage}{.33\linewidth}
\centering
\begin{tabular}{cc}
\toprule
Slot Value & Count \\
\midrule
   first  & 21\\
   5th  & 11\\
   second & 8 \\
   3rd & 8 \\
   4th & 7 \\
\bottomrule
\end{tabular}
\caption{ORDINAL slot values.}
\end{minipage}
\end{table}

\begin{table}[H]
\begin{minipage}{.5\linewidth}
\centering
\begin{tabular}{cc}
\toprule
Slot Value & Count \\
\midrule
   mastercard  &  58\\
   401k  & 49\\
   bank of america & 39 \\
   chase & 39 \\
   american express & 37 \\
\bottomrule

\end{tabular}
\caption{ORG slot values.}
\end{minipage}
\begin{minipage}{.5\linewidth}
\centering
\begin{tabular}{cc}
\toprule
Slot Value & Count \\
\midrule
   delta  & 19\\
   africa  & 14\\
   europe & 7 \\
   asia & 6 \\
   the grand canyon & 3 \\
\bottomrule
\end{tabular}
\caption{LOC slot values.}
\end{minipage}

\end{table}

\section{Dialogue Embedding Evaluations}
\label{sec:eval}

In line with prior publications on dialogue sentence embeddings~\cite{zhou2022learning,wu2020todbert,liu2021dialoguecse,burdisso-etal-2024-dialog2flow}, we do not evaluate on general benchmarks such as~\cite{muennighoff2023mteb, senteval-2018, thakur2021beir} or STS~\cite{cer-etal-2017-semeval}. This is in contrast to general sentence embeddings~\cite{zhang-etal-2023-contrastive-learning, cheng-etal-2023-improving, wang2024text, wang2024improving, izacard2022unsupervised}. The reason being:

"Here we do not adopt the standard semantic textual similarity (STS) task for two reasons: (1) The sentence embedding performance varies greatly as the domain of the training data changes. As a dialogue dataset is always about several certain domains, evaluating on the STS benchmark may mislead the evaluation of the model. (2) The dialogue-based sentence embeddings focus on context-aware rather than context-free semantic meanings, which may not be suitable to be evaluated through the context-free benchmarks."~\cite{liu2021dialoguecse}

\section{Configuration}
\label{sec:config}

\begin{table}[H]
    \centering
    \begin{tabular}{cccc}
        \toprule
        Src. Data & Test & Slots & Intents \\
        \midrule
        ATIS & 893 & 129 & 26 \\
        HWU64 & 1076 & 54 & 64 \\
        SNIPS & 700 & 53 & 7\\
        MASSIVE & 2974 & 55 & 60\\
        CLINC150 & 5500 & 17 (aug.) & 150\\
        \bottomrule
    
    \end{tabular}
    \caption{Statistics of original source dialogue datasets. Note that the slot count in CLINC150 is from our slot-filling augmentation.}
    \label{tab:dataset}
\end{table}

We perform transfer learning on top of the SimCSE~\cite{simcse-2021} BERT-base (110M params) model (unsup-simcse-bert-base-uncased), as our purpose is to evaluate dialogue-specific effects. Table~\ref{tab:main_exp} uses augmented data from all sources (SNIPS / ATIS / MASSIVE / HWU64 / CLINC150) while other tables experiment with single-source data. For "ours" models in Table~\ref{tab:main_exp}, transfer learning is performed on corresponding checkpoints from each publication using our augmented data and SimCSE loss ($L^{u}$-only). For baselines in other tables, we experiment with the same BERT-base variants. We utilize a low learning rate of $1e-8$ and train for 2 epochs for all models. The only exception is considering 8 epochs for single-source MLP pooled models to learn the $W_A$ properly (Fig.~\ref{fig:epoch}). Batch size 16. We experiment with $\lambda^t \in \{1.0, 0.0\}$, $\lambda^u \in \{1.0, 0.0\}$ and $\lambda^\textit{pair} \in \{0.5, 0.0\}$ and $\lambda^\textit{comp} \in \{0.1, 0.2, 0.5\}$. We select best $\lambda^\textit{comp}$ per configuration according to the validation set. We perform experiments on RTX 3090. We only use NLP research tools as they are intended for research purposes. Public dialogue datasets do not contain identifiable information.

\begin{figure}[H]
    \centering
    \includegraphics[width=.5\columnwidth]{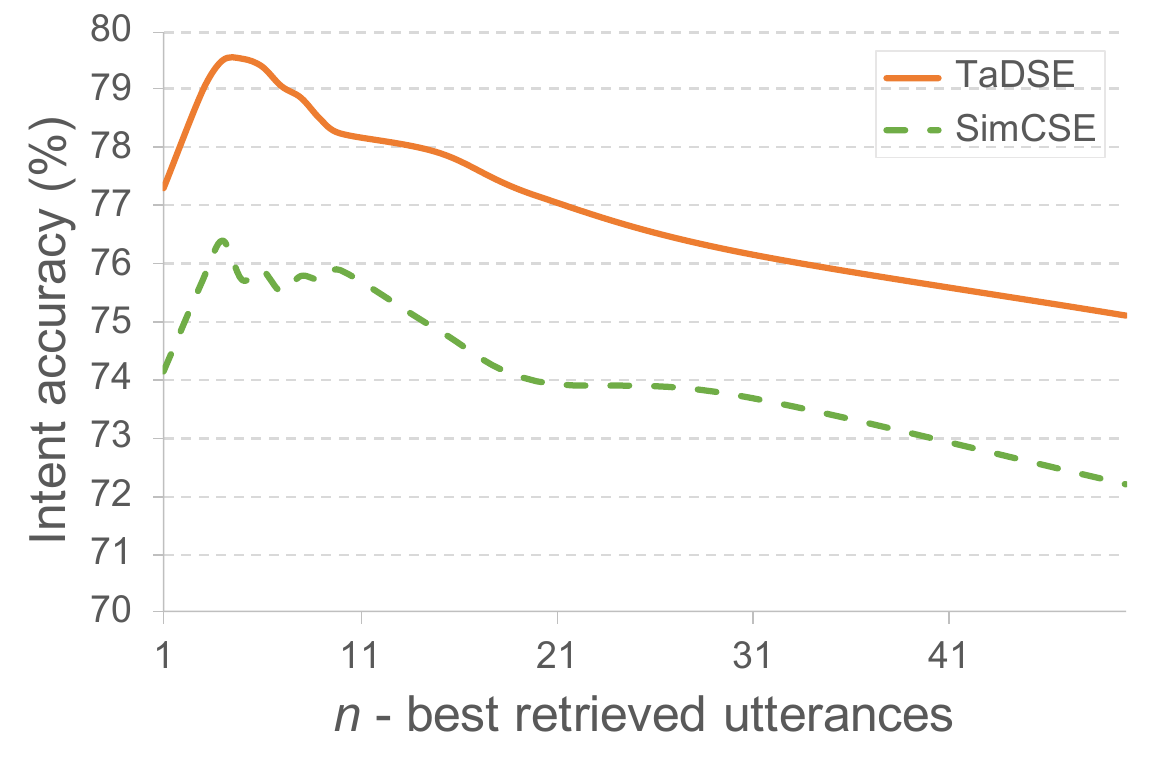}
    \caption{MASSIVE performance with TaDSE and SimCSE. The horizontal axis is the $k$ value in kNN.}
    \label{fig:intent_acc}
\end{figure}

We evaluate with kNN to select most relevant reference vector to extract intent information ($k=1$, experiment in Fig.~\ref{fig:intent_acc}). We assert that our choice of evaluation method (kNN) emphasizes the local structure of the representation space in contrast to the global structure. This is fitting for our approach as a template representation may reside close to utterance representation, affecting local structure more. We evaluate with full utterance/template training set representations and we include CLINC150 OOS labels. 

\section{Training Stability}
\label{sec:train_stable}

\begin{figure}[H]
     \centering
     \includegraphics[width=0.5\columnwidth]{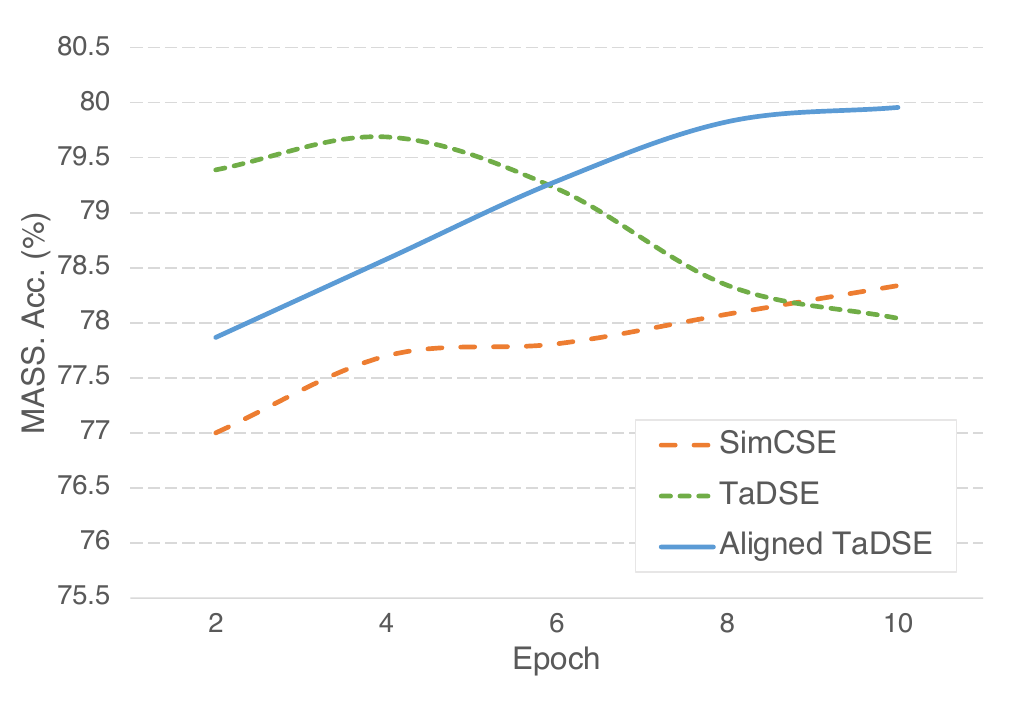}
     \caption{Performance on test set during 10 training epochs. 'Aligned' model is the 'w/ MLP' variant.}
 \label{fig:epoch}
\end{figure}

We report separate evaluations with the inclusion of MLP layer $W_A$, which improve performance on certain datasets (Table~\ref{tab:main_exp}, $L^{t'}$ in Table~\ref{tab:pairwise_exp}). As the MLP layer needs to be trained from scratch, we empirically require more training epochs than non-MLP TaDSE in our experiments (Fig.~\ref{fig:epoch}, Section~\ref{sec:config}). In contrast, non-MLP TaDSE performance is optimal at a lower epoch.

\section{Uniformity / Alignment Definition}
\label{sec:uniform_align_def}

We compute uniformity/alignment per test set of each source data and define $p_\textit{pos}$ as representations within the same label, and $p_\textit{data}$ as the sentences from each original non-augmented source dataset.

Uniformity is a measurement of the degree of uniformness of the representations :

\begin{equation}
    \ell_{\textit{uniform}}\triangleq\log \underset{~~~x, y\stackrel{i.i.d.}{\sim} p_{\textit{data}}}{\mathbb{E}}   e^{-2\Vert f(x)-f(y) \Vert_2^2}
    \label{eq:uniformity}
\end{equation}

Alignment measures the distance between positive representations :

\begin{equation}
    \ell_{\textit{align}}\triangleq \underset{(x, x^+)\sim p_{\textit{pos}}}{\mathbb{E}} \Vert f(x) - f(x^+) \Vert_2^2
    \label{eq:alignment}
\end{equation} 

\section{ATIS T-SNE Diagrams}
\label{sec:atis_tsne}

\begin{figure}[H]
\centering
    \advance\rightskip0.5cm
\includegraphics[width=.5\columnwidth]{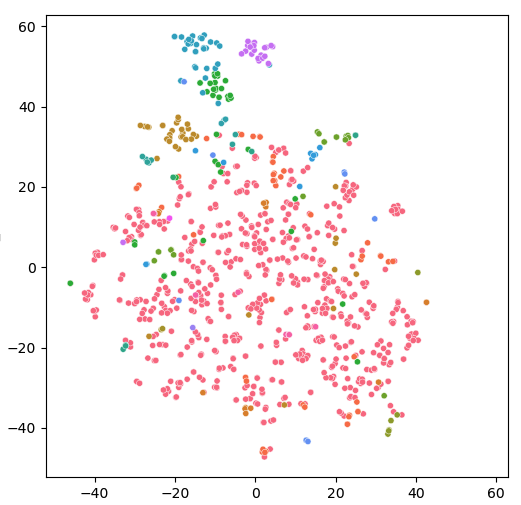}
\caption{T-SNE diagram for ATIS representation hyperspace from SimCSE model, trained with our data.}
\label{fig:atis_base}
\end{figure}

\begin{figure}[H]
\centering
    \advance\rightskip0.5cm
\includegraphics[width=.5\columnwidth]{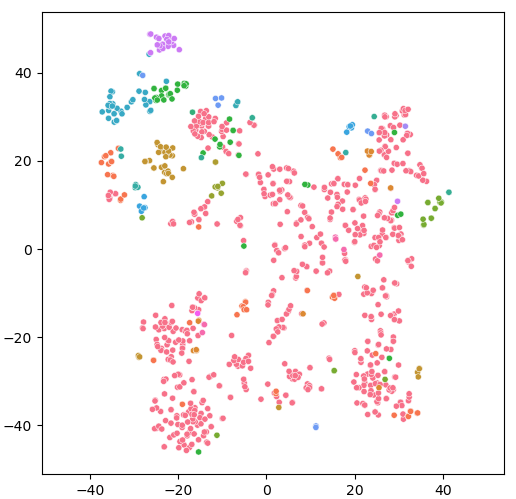}
\caption{T-SNE diagram for ATIS representation hyperspace from TaDSE model, trained with our data.}
\label{fig:atis_train}
\end{figure}

\begin{figure}[H]
\centering
    \advance\rightskip0.5cm
\includegraphics[width=.5\columnwidth]{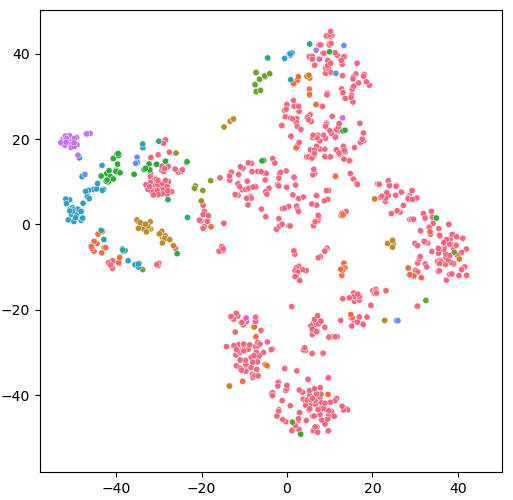}
\caption{T-SNE diagram for ATIS representation hyperspace from our optimal TaDSE model ($\lambda^\textit{comp} = 0.2$).}
\label{fig:atis_optimal}
\end{figure}

\end{document}